\documentclass[runningheads]{paper305}
\usepackage[rgb,x11names]{xcolor}
\usepackage{amssymb}
\usepackage{amsmath} 
\usepackage{subfig}
\usepackage{hyperref}
\hypersetup{colorlinks=true,citecolor=blue}
\usepackage{graphicx}
\usepackage[]{algorithm2e}
\usepackage{mathrsfs}
\usepackage{bbding}
\usepackage{rotating}
\usepackage{multirow}
\usepackage{marginnote}
\usepackage{overpic}
\usepackage{colortbl}
\usepackage[labelfont=bf]{caption}
\usepackage{bbding}

\newcommand{\textBC}[2]{\textbf{\textcolor{#1}{#2}}}
\definecolor{mygray}{gray}{.92}

\def\ourmodel{\textit{MSNet}}
\usepackage{graphicx}
\begin{document}
%
\title{Automatic Polyp Segmentation via\\ Multi-scale Subtraction Network}
%
%

\author{Xiaoqi Zhao\inst{1}, Lihe Zhang\inst{1}\thanks{Corresponding author.}, and Huchuan Lu\inst{1,2}}
\index{Zhao, Xiaoqi}
\index{Zhang, Lihe}
\index{Lu, Huchuan}

\authorrunning{Zhao et al.}
%
\institute{Dalian University of Technology, China 
\and
Peng Cheng Laboratory 
\\
\email{\ zxq@mail.dlut.edu.cn, \{zhanglihe,lhchuan\}@dlut.edu.cn}
}
\maketitle              
\begin{abstract}
More than 90\% of colorectal cancer is gradually transformed from colorectal polyps. In clinical practice, precise polyp segmentation provides important information in the early detection of colorectal cancer. Therefore, automatic polyp segmentation techniques are of great importance for both patients and doctors. Most existing 
methods are based on U-shape structure and 
use element-wise addition or concatenation to fuse different level features progressively in decoder. However, both the two operations easily generate plenty of redundant information, which will weaken the complementarity between different level features, resulting in inaccurate localization and blurred edges of polyps. To address this challenge, we propose a multi-scale subtraction network (MSNet) to segment polyp from colonoscopy image. Specifically, we first 
design a subtraction unit (SU) to produce the difference features between adjacent levels in encoder. Then, 
we pyramidally equip the SUs at different levels with varying receptive fields, thereby obtaining rich multi-scale difference information. In addition, 
we build a training-free network ``LossNet'' to comprehensively supervise the polyp-aware features from bottom layer to top layer, which drives the MSNet to capture the detailed and structural cues simultaneously. Extensive experiments on five benchmark datasets demonstrate that our MSNet performs favorably against most state-of-the-art methods under different evaluation metrics. Furthermore, MSNet runs at a real-time speed of $\sim$70fps when processing a $352 \times 352$ image. The source code will be publicly available at \url{https://github.com/Xiaoqi-Zhao-DLUT/MSNet}.
\end{abstract}

\section{Introduction}
According to GLOBOCAN 2020 data, colorectal cancer is the third most common cancer worldwide and the second most common cause of death. It usually begins as small, noncancerous (benign) clumps of cells called polyps that form on the inside of the colon. Over time some of these polyps can become colon cancers. Therefore, the best way of preventing colon cancer is to identify and remove polyps before they turn into cancer. At present, colonoscopy is the most commonly used means of examination, but this process involves manual and expensive labor, not to mention its high misdiagnosis rate. Hence, automatic and accurate polyp segmentation is of great practical significance. 

The automatic polyp segmentation has gradually evolved from the traditional methods~\cite{tajbakhsh2015automatic} based on manually designed features to the deep learning methods~\cite{yu2016integrating,zhang2018polyp,pnsnet}.  Although these methods have made progress in clinical, they are limited by box-level prediction results, thus failing to capture the shape and contour of polyps. To address this issue, Brandao \textit{et al.}~\cite{brandao2017fully} utilize the FCN~\cite{FCN} to segment polyps by a pixel-level prediction. Akbari \textit{et al.}~\cite{akbari2018polyp} also use FCN-based segmentation network and combine the patch selection mechanism to improve the accuracy of polyp segmentation. However, FCN-based methods rely on low-resolution features to generate the final prediction, resulting in rough segmentation results and fuzzy boundaries. 

In recent years, U-shape structures~\cite{FPN,UNet,MRNet} have received considerable attention due to their abilities of utilizing multi-level information to reconstruct high-resolution feature maps. Many polyp segmentation networks~\cite{UNet,UNet++,SFA,PraNet} adopt the U-shape  architecture. In UNet~\cite{UNet}, the up-sampled feature maps are concatenated with feature maps skipped from the encoder and convolutions and non-linearities are added between up-sampling steps. UNet++~\cite{UNet++} uses nested and dense skip connections to reduce the  semantic  gap  between  the feature  maps  of  encoder  and  decoder.
Later, ResUNet++~\cite{ResUnet++} combines many advanced techniques such as residual computation~\cite{ResNet}, squeeze and excitation~\cite{SENet}, atrous spatial pyramidal pooling~\cite{Deeplab}, and  attention mechanism to further improve performance. Recent works, SFA~\cite{SFA} and PraNet~\cite{PraNet}, focus on recovering the sharp boundary between a polyp and its surrounding mucosa. The former proposes a selective feature aggregation structure and a boundary-sensitive loss function under a shared encoder and two mutually constrained decoders. The latter utilizes reverse attention module to establish the relationship between region and boundary cues. 
%

Generally speaking, different level features in encoder have different characteristics. High-level ones have more semantic information which helps localize the objects, while low-level ones have more detailed information which can capture the subtle boundaries of objects. The  decoder leverages the level-specific and cross-level characteristics to generate the final high-resolution prediction. Nevertheless, the aforementioned methods directly use an element-wise addition or concatenation to fuse any two level features from the encoder and transmit them to the decoder. These simple operations do not pay more attention to differential information between different levels. 
This drawback not only generates redundant information to dilute the really useful features but also weakens the characteristics of level-specific features, which results in that the network can not balance 
accurate polyp localization and subtle boundary refinement.

\begin{figure}[t]
\centering
\includegraphics[width=0.6\linewidth]{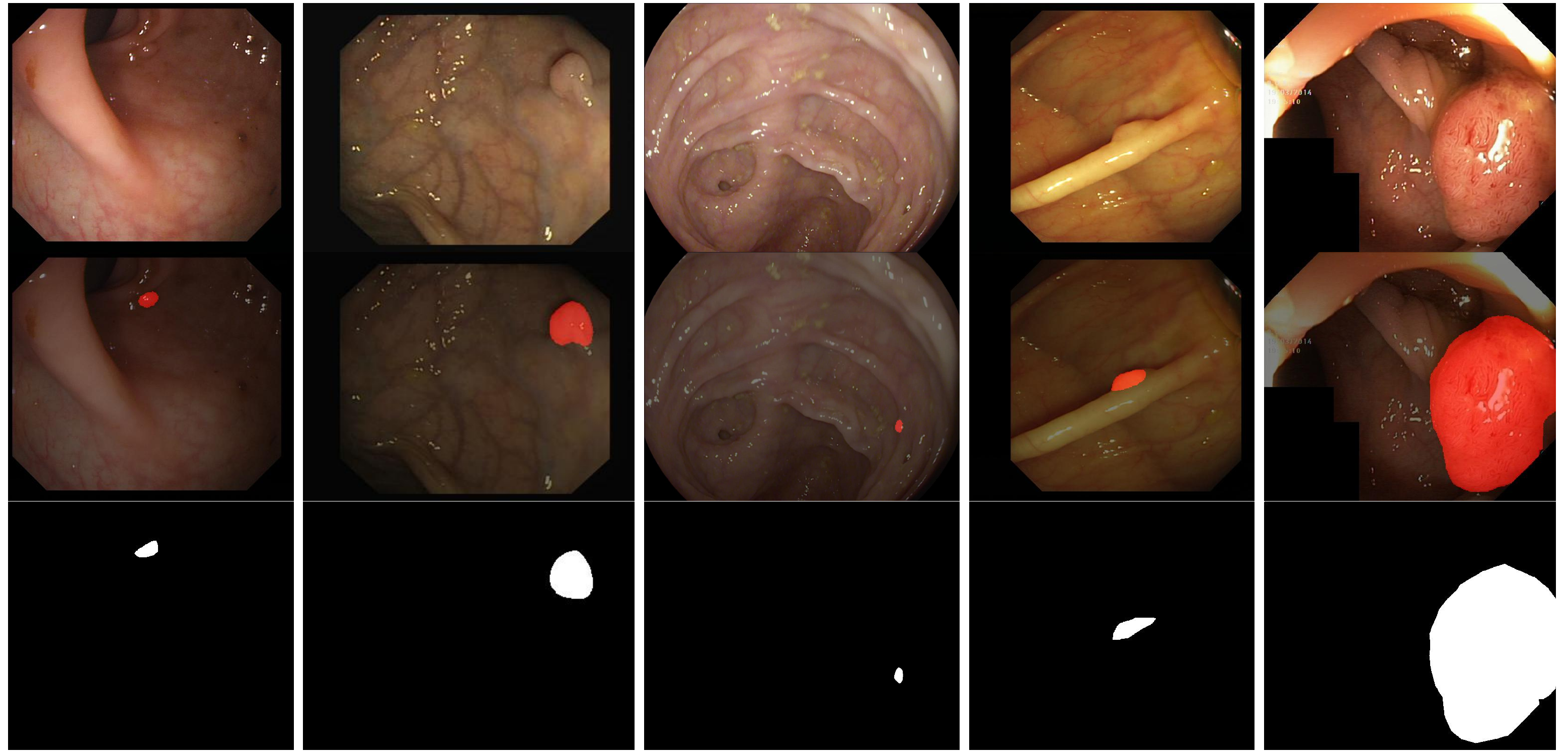}
\caption{Visualization of RGB color slices, our prediction and gold standard.}
\label{fig:rgb_visual_detector}
\end{figure}

In this paper,  we propose a novel multi-scale subtraction network (MSNet) for the polyp segmentation task. 
We first 
design a subtraction unit (SU) and apply it to each pair of adjacent level features. 
To address the scale diversity of polyps, we pyramidally concatenate multiple SUs to capture the large-span cross-level information. 
Then, we aggregate level-specific features and multi-path cross-level differential features and then generate the final prediction in decoder.
Moreover, we propose a LossNet to automatically supervise the extracted feature maps from bottom layer to top layer, which can optimize the segmentation from detail to structure with a simple L2-loss function.
Our main contributions can be summarized as follows:
\begin{itemize} 
\item  We propose a novel multi-scale subtraction network for automatic polyp segmentation. With multi-level and multi-stage cascaded subtraction operations, the complementary information from lower order to higher order among different levels can be effectively obtained, thereby comprehensively enhancing the perception of polyp areas.
\item  We build a general training-free loss network to implement the detail-to-structure supervision in the feature levels, which provides important supplement to the loss design based on the prediction itself.
\item The proposed MSNet can accurately segment polyps as shown in Fig.~\ref{fig:rgb_visual_detector}. Extensive experiments demonstrate that our MSNet advances the state-of-the-art methods by a large margin under different evaluation metrics on five challenging datasets, with a real-time inference speed of $\sim$70fps.
\end{itemize}

\section{Method}
The MSNet architecture is shown in Fig.~\ref{fig:Pipeline}, in which there are five encoder blocks ($\mathbf{E}^i$, $i \in \left \{1, 2, 3, 4, 5 \right \}$), a multi-scale subtraction module and four decoder blocks ($\mathbf{D}^i$, $i \in \left \{1, 2, 3, 4 \right \}$).  Following the PraNet~\cite{PraNet}, we adopt the Res2Net-50 as the backbone to extract five levels of features. First, we separately adopt a $3 \times 3$ convolution for feature maps of each encoder block to reduce the channel to $64$, which can decrease the number of parameters for subsequent operations. Next, these different level features are fed into the multi-scale subtraction module and output five complementarity 
enhanced features (${CE}^i$, $i \in \left \{1, 2, 3, 4, 5 \right \}$). Finally, each ${CE}^i$ progressively participates in the decoder and generate the final prediction. In the training phase, both the prediction and ground truth are fed into the LossNet to achieve  supervision. We describe the multi-scale subtraction module in Sec.~\ref{sec:msm} and give the details of LossNet in Sec.~\ref{sec:lossnet}.
\subsection{Multi-scale Subtraction Module}\label{sec:msm}
\begin{figure}[t]
\centering
\includegraphics[width=0.8\linewidth]{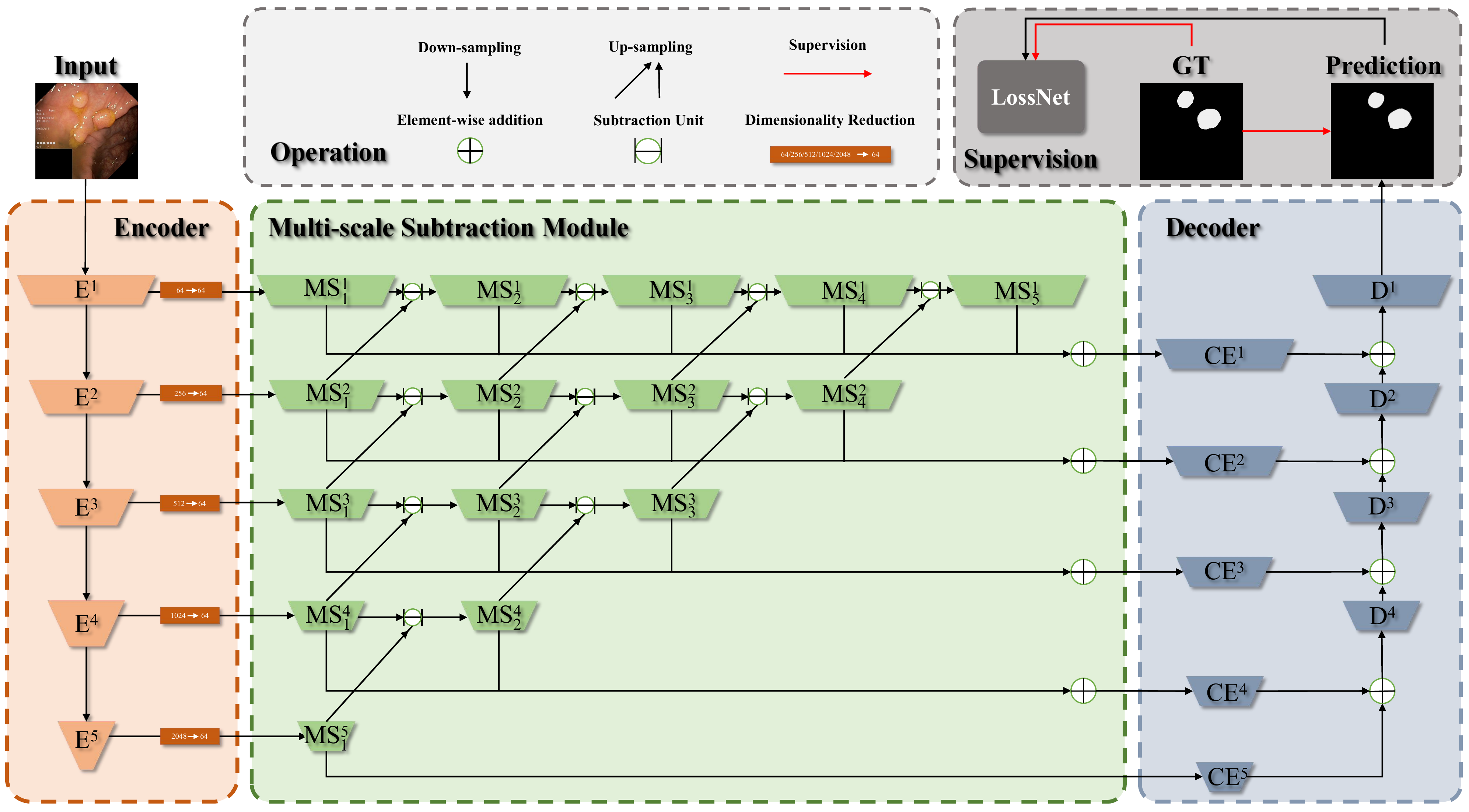}
\caption{Overview of the proposed MSNet.}
\label{fig:Pipeline}
\end{figure}

We use  $F_{A}$ and $F_{B}$ to represent adjacent level feature maps. They all have been activated by the ReLU operation. We  define a basic subtraction unit (SU):
  \begin{equation}\label{equ:2}
 \begin{split}
     SU = Conv(\vert F_{A} \ominus F_{B} \vert ),
 \end{split}
\end{equation}
 where $\ominus$ is the element-wise subtraction operation, $ \vert \cdot \vert$ calculates the absolute value and $Conv(\cdot)$ denotes the convolution layer. 
 The SU unit can capture the complementary information of $F_{A}$ and $F_{B}$ and highlight their differences, thereby providing richer information for the decoder.

To obtain higher-order complementary information across multiple feature levels,  
we horizontally and vertically concatenate multiple SUs to calculate a series of differential features with different orders and receptive fields. The detail of the multi-scale subtraction module can be found in Fig.~\ref{fig:Pipeline}. We aggregate the level-specific feature ($MS^{i}_{1}$) and cross-level differential features ($MS^{i}_{n \neq 1}$) between the corresponding level and any other levels  
to generate complementarity enhanced feature ($CE^{i}$). This process can be formulated as follows:
\begin{equation}\label{equ:2}
 \begin{split}
     CE^{i} = Conv(\sum_{n=1}^{6-i}MS^{i}_{n} ) \quad i=1, 2, 3, 4, 5.
 \end{split}
\end{equation}
Finally, all $CE^{i}$ participate in decoding and then the polyp region is segmented.  
\subsection{LossNet}\label{sec:lossnet}
\begin{figure}[t]
\centering
\includegraphics[width=0.7\linewidth]{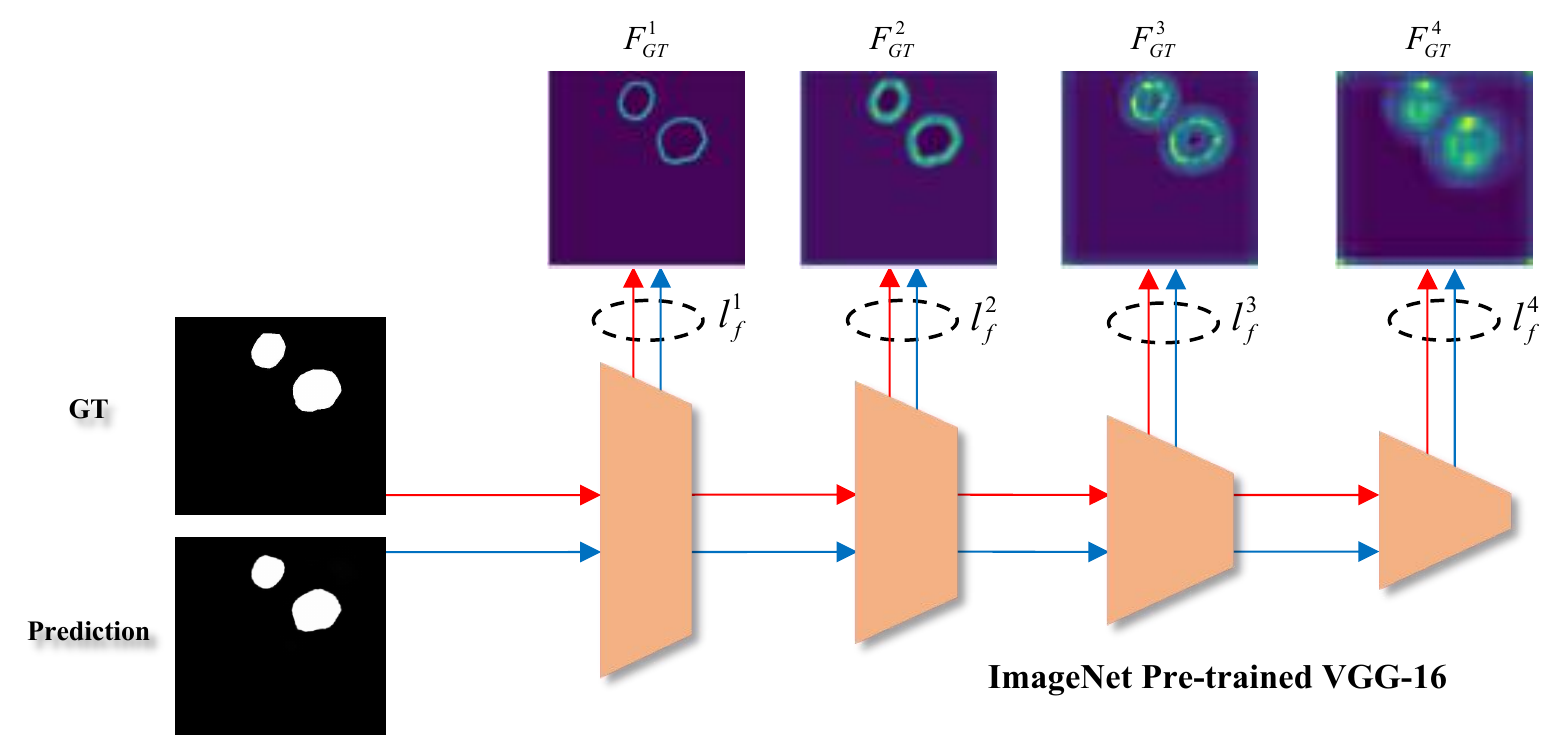}
\caption{Illustration of LossNet.}
\label{fig:P_loss}
\end{figure}
In the proposed model, the total training loss can be written as:
\begin{equation}\label{equ:3}
 \begin{split}
    \mathcal{L}_{total}=\mathcal{L}_{IoU}^w +\mathcal{L}_{BCE}^w + \mathcal{L}_{f},
 \end{split}
\end{equation}
 where $\mathcal{L}_{IoU}^w$ and $\mathcal{L}_{BCE}^w$ represent the weighted IoU loss and binary cross entropy (BCE) loss which have been widely adopted in segmentation tasks. We use the same definitions as in ~\cite{PraNet,wei2019f3net,qin2019basnet} and their effectiveness has been validated in theses works. Different from them, we extra use a LossNet to further optimize the segmentation from detail to structure. Specifically, we use an ImageNet pre-trained classification network, such as VGG-16, to extract the multi-scale features of the prediction and ground truth, respectively. Then, their feature difference is computed as loss $\mathcal{L}_{f}$: 
\begin{equation}\label{equ:3}
 \begin{split}
    \mathcal{L}_{f} = {l}_{f}^{1} + {l}_{f}^{2} + {l}_{f}^{3} + {l}_{f}^{4}.
 \end{split}
\end{equation}
Let ${F}_{P}^{i}$ and ${F}_{G}^{i}$ separately represent the $i$-th level feature maps extracted from the prediction and ground truth. The ${l}_{f}^{i}$ is calculated as their Euclidean distance (L2-Loss), which is supervised at the pixel level:
\begin{equation}\label{equ:3}
 \begin{split}
    {l}_{f}^{i} = \vert\vert {F}_{P}^{i} -  {F}_{G}^{i} \vert\vert_{2}, \quad i=1, 2, 3, 4.
 \end{split}
\end{equation}
The structure of LossNet is shown in Fig~\ref{fig:P_loss}. It can be seen that the low-level feature maps contain rich boundary information and the high-level ones depict  location information. Thus, the LossNet can generate comprehensive supervision in the feature levels. 

\section{Experiments}
\subsection{Datasets}
We evaluate the proposed model on five benchmark datasets: CVC-ColonDB~\cite{CVC-ColonDB}, ETIS~\cite{ETIS}, Kvasir~\cite{Kvasir}, CVC-T~\cite{CVC-T} and CVC-ClinicDB~\cite{CVC-ClinicDB}. We adopt the same training set as the latest polyp segmentation method~\cite{PraNet}, that is,  $900$ samples from the Kvasir and $550$ samples from the CVC-ClinicDB are used for training. 
The remaining images and other three datasets are used for testing.
\subsection{Evaluation Metrics}
We adopt several widely used metrics for quantitative evaluation: mean Dice, mean IoU, the weighted F-measure ($F_{\beta}^{w}$)~\cite{Fwb}, mean absolute error (MAE), the recently released S-measure ($S_{\alpha}$)~\cite{S-m} and E-measure ($E_\phi^{max}$)~\cite{Em} scores. The lower value is better for the MAE and the higher is better for others.
\subsection{Implementation Details}
Our model is implemented based on the PyTorch framework and trained on a single 2080Ti GPU for $50$ epochs with mini-batch size $16$. We resize the inputs to $352 \times 352$ and employ a  general multi-scale training strategy as the PraNet~\cite{PraNet}.  Random horizontally flipping and random rotate data augmentation are used to avoid overfitting. For the optimizer, we adopt the stochastic gradient descent (SGD). The momentum and weight decay are set as $0.9$ and $0.0005$, respectively.  Maximum learning rate is set to $0.005$ for backbone and $0.05$ for other parts. Warm-up and linear decay strategies  are  used to  adjust  the  learning  rate. 
\subsection{Comparisons with State-of-the-art}
We compare our MSNet with U-Net~\cite{UNet}, U-Net++~\cite{UNet++}, SFA~\cite{SFA} and PraNet~\cite{PraNet}.  To be fair,  the predictions of these competitors are directly provided by their respective authors or computed by their released codes.

\textbf{Quantitative Evaluation.}
Tab.~\ref{tab:performance} shows performance comparisons in terms of six metrics.  It can be seen that our MSNet outperforms other approaches across all datasets. In particular, MSNet achieves a predominant performance on the CVC-ColonDB and ETIS datasets. Compared to the second best method (PraNet), our method achieves an important improvement on the challenging ETIS of $14.1\%$, $15.3\%$, $13.0\%$, $6.2\%$, $4.8\%$ and $35.5\%$  in terms of mDice, mIoU, $F_\beta^w$, $S_{\alpha}$, $E_\phi^{max}$ and MAE, respectively. In addition, Tab.~\ref{tab:speed} lists the model average speed of different methods. Our model runs at a real-time speed of $\sim$70fps that is the fastest one among these state-of-art methods.

\begin{figure}[!h]
\centering
\includegraphics[width=0.6\linewidth]{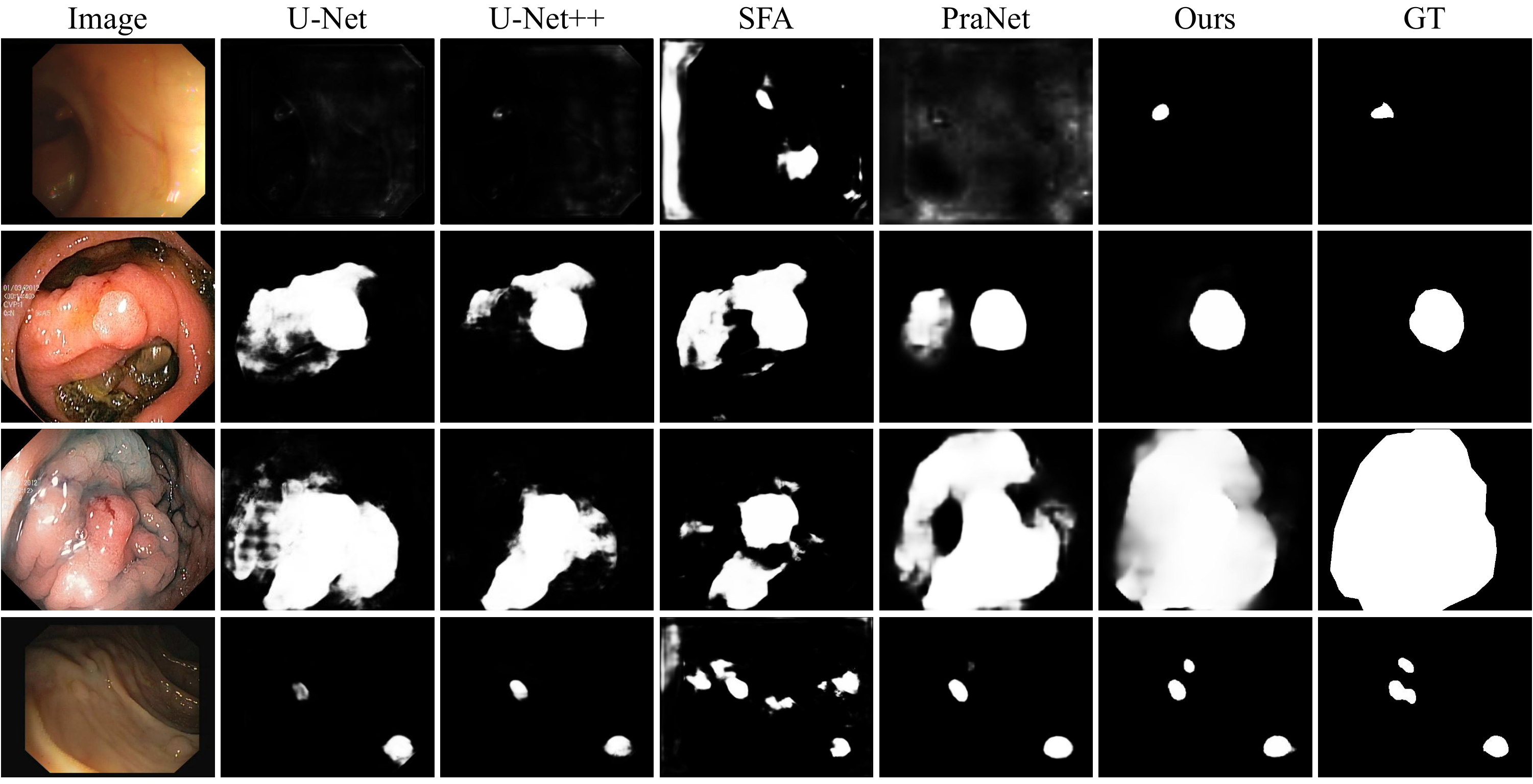}
\caption{Visual comparison of different methods.}
\label{fig:visual_comparison}
\end{figure}

\textbf{Qualitative Evaluation.}
Fig.~\ref{fig:visual_comparison} illustrates visual comparison with other approaches. It can be seen that the proposed method has good detection performance for small, medium, and large scale polyps (see the $1^{st}$ - $3^{th}$ rows). Moreover, for the images with multiple polyps, our method can accurately detect them and capture more details (see the $4^{th}$ rows).
\begin{table*}[t]
  \centering
  \scriptsize
  \renewcommand{\arraystretch}{1.1}
  \setlength\tabcolsep{5pt}
  \caption{ Quantitative comparison. $\uparrow$ and $ \downarrow$ indicate that the larger and smaller scores are better, respectively. The best results are shown in $\textBC{red}{red}$ .
  }\label{tab:performance}
  \resizebox{0.8\columnwidth}{!}
  {
  \begin{tabular}{cr||cccccccc}
  \hline
  \rowcolor{mygray}
  &Methods & mDice $\uparrow$ & mIoU $\uparrow$  &  $F_\beta^w$  $\uparrow$& $S_{\alpha}$ $\uparrow$&$E_\phi^{max}$ $\uparrow$ & MAE $\downarrow$\\
  \hline
  \multirow{5}{*}{\begin{sideways}ColonDB\end{sideways}} & 
  U-Net(MICCAI'15)~\cite{UNet}  & 0.519 & 0.449 & 0.498 & 0.711 & 0.763 & 0.061 \\
  &U-Net++(TMI'19)~\cite{UNet++} & 0.490 & 0.413 & 0.467 & 0.691 & 0.762 & 0.064 \\
  &SFA (MICCAI'19)~\cite{SFA} & 0.467 & 0.351 & 0.379 & 0.634 & 0.648 & 0.094\\
  &PraNet (MICCAI'20)~\cite{PraNet} & 0.716 & 0.645 & 0.699 & 0.820 & 0.847 & 0.043\\
  &\textbf{\ourmodel~(Ours)}  & \textBC{red}{0.755} & \textBC{red}{0.678} & \textBC{red}{0.737} & \textBC{red}{0.836} & \textBC{red}{0.883} & \textBC{red}{0.041}\\
  \hline
  \hline
  \multirow{5}{*}{\begin{sideways}ETIS\end{sideways}} &
  U-Net (MICCAI'15)~\cite{UNet}  & 0.406 & 0.343 & 0.366 & 0.682 & 0.645 & 0.036 \\
  &U-Net++ (TMI'19)~\cite{UNet++}  & 0.413 & 0.342 & 0.390 & 0.681 & 0.704 & 0.035 \\
  &SFA (MICCAI'19)~\cite{SFA} & 0.297 & 0.219 & 0.231 & 0.557 & 0.515 & 0.109\\
  &PraNet (MICCAI'20)~\cite{PraNet} & 0.630 & 0.576 & 0.600 & 0.791 & 0.792 & 0.031\\
  &\textbf{\ourmodel~(Ours)}  & \textBC{red}{0.719} & \textBC{red}{0.664} & \textBC{red}{0.678} & \textBC{red}{0.840}  & \textBC{red}{0.830} & \textBC{red}{0.020}\\
  \hline
  \hline
  \multirow{5}{*}{\begin{sideways}Kvasir\end{sideways}} &
  U-Net (MICCAI'15)~\cite{UNet}  & 0.821 & 0.756 & 0.794 & 0.858 & 0.901 & 0.055\\
  &U-Net++ (TMI'19)~\cite{UNet++}   & 0.824 & 0.753 &0.808& 0.862 & 0.907 &  0.048\\
  &SFA (MICCAI'19)~\cite{SFA} & 0.725 & 0.619 & 0.670 & 0.782 & 0.828 & 0.075\\
  &PraNet (MICCAI'20)~\cite{PraNet} & 0.901 & 0.848 & 0.885 & 0.915 & 0.943 & 0.030\\
  &\textbf{\ourmodel~(Ours)}  & \textBC{red}{0.907} & \textBC{red}{0.862} & \textBC{red}{0.893} & \textBC{red}{0.922} & \textBC{red}{0.944} & \textBC{red}{0.028} \\
  \hline
  \hline
  \multirow{5}{*}{\begin{sideways}CVC-T\end{sideways}} &
  U-Net (MICCAI'15)~\cite{UNet}  & 0.717 & 0.639 & 0.684 & 0.842 & 0.867 & 0.022\\
  &U-Net++ (TMI'19)~\cite{UNet++}   & 0.714 & 0.636 &0.687& 0.838 & 0.884 &  0.018\\
  &SFA (MICCAI'19)~\cite{SFA} & 0.465 & 0.332 & 0.341 & 0.640 & 0.604 & 0.065\\
  &PraNet (MICCAI'20)~\cite{PraNet} & \textBC{red}{0.873} & 0.804 & 0.843 & 0.924 & 0.938 & \textBC{red}{0.010}\\
  &\textbf{\ourmodel~(Ours)}  & 0.869 & \textBC{red}{0.807} & \textBC{red}{0.849} & \textBC{red}{0.925} & \textBC{red}{0.943} & \textBC{red}{0.010} \\
  \hline
  \hline
  \multirow{5}{*}{\begin{sideways}ClinicDB\end{sideways}} &
  U-Net (MICCAI'15)~\cite{UNet}  & 0.824 & 0.767 & 0.811 & 0.889 & 0.917 & 0.019\\
  &U-Net++ (TMI'19)~\cite{UNet++}   & 0.797 & 0.741 &0.785& 0.872 & 0.898 &  0.022\\
  &SFA (MICCAI'19)~\cite{SFA} & 0.698 & 0.615 & 0.647 & 0.793 & 0.816 & 0.042\\
  &PraNet (MICCAI'20)~\cite{PraNet} & 0.902 & 0.858 & 0.896 & 0.935 & 0.958 & 0.009\\
  &\textbf{\ourmodel~(Ours)}  & \textBC{red}{0.921} & \textBC{red}{0.879} & \textBC{red}{0.914} & \textBC{red}{0.941} &\textBC{red}{0.972} &\textBC{red}{0.008} \\
  \hline
  \end{tabular}
  }
\end{table*}

\begin{table*}
  \centering
  \scriptsize
  \renewcommand{\arraystretch}{1.1}
  \setlength\tabcolsep{5pt}
  \caption{ The average speed of different methods.
  }\label{tab:speed}
  \begin{tabular}{c||c|c|c|c|c}
  \hline
  \rowcolor{mygray}
  Methods & U-Net  & U-Net++  &  SFA & PraNet & \textbf{\ourmodel~(Ours)} \\
  \hline
  Average speed & $\sim$8fps  & $\sim$7fps  & $\sim$40fps & $\sim$50fps & \textBC{red}{$\sim$70fps} \\
  \hline
\end{tabular}
\end{table*}
\begin{table*}[!ht]
  \centering
  \scriptsize
  \renewcommand{\arraystretch}{1.1}
  \setlength\tabcolsep{5pt}
    \caption{Ablation study on the CVC-ColonDB and ETIS datasets.}\label{tab:ablation_study}
     \resizebox{0.7\columnwidth}{!}
     {
	\begin{tabular}{c||cccc|cccc}
	 \hline
	 \rowcolor{mygray}
	 & \multicolumn{4}{c|}{ColonDB} & \multicolumn{4}{c}{ETIS} \\
	\rowcolor{mygray}
	 Metric & mDice  & mIoU  & $F_\beta^w$ &$E_\phi^{max}$ & mDice  & mIoU  & $F_\beta^w$ &$E_\phi^{max}$ \\
	\hline
	baseline ($MS_{1}^{i}$) &{0.678} &{0.607} &{0.659} & {0.825} & {0.588} &0.549 &{0.532} & {0.707} \\
	\hline
	+ $MS_{2}^{i}$ &{0.731} &{0.652} &{0.703} & {0.861} & {0.642} &0.579 &{0.586} & {0.745} \\
	\hline
	+ $MS_{3}^{i}$ &{0.733} &{0.659} &{0.712} & {0.861} & {0.642} &0.580 &{0.581} & {0.745} \\
	\hline
	+ $MS_{4}^{i}$ &{0.750} &{0.676} &{0.729} & {0.872} & {0.643} &0.580 &{0.585} & {0.757} \\
	\hline
	+ $MS_{5}^{i}$ &{0.749} &{0.676} &{0.729} & {0.878} & {0.643} &0.582 &{0.600} & {0.787} \\
	\hline
	+ $\mathcal{L}_{f}$ &{0.755} &{0.678} &{0.737} & {0.883} & {0.719} &0.664 &{0.678} & {0.830} \\
	\hline
	Replace $MS$ with $MA$ &{0.697} &{0.630} &{0.676} & {0.839} & {0.680} &0.621 &{0.636} & {0.820} \\
	\hline
	\end{tabular}
	}
\end{table*}
\begin{figure}
\centering
\includegraphics[width=0.8\linewidth]{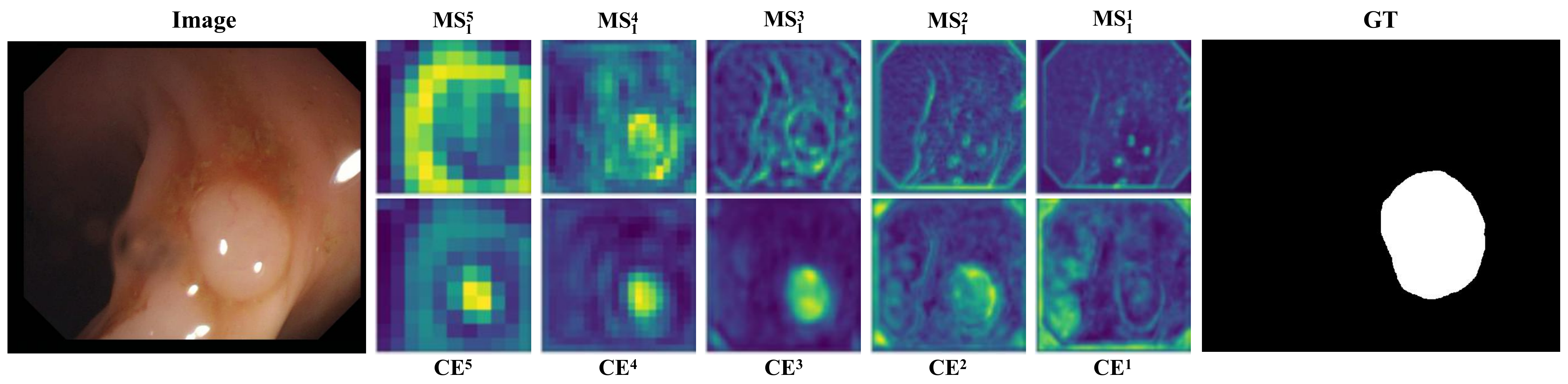}
\caption{Visual comparison between encoder features and the cross-level complementarity enhanced features.}
\label{fig:ablation_study_visual_encoder_transition}
\end{figure}

\subsection{Ablation Study}
We take the common FPN network as the baseline to analyze the contribution of each component. The results are shown in Tab~\ref{tab:ablation_study}. These defined feature symbols are the same as those in Fig~\ref{fig:Pipeline}.
First, we apply the subtraction module to the baseline to get a  series of $MS_{i}^{2}$  features to participate in the feature aggregation calculated by Equ.~\ref{equ:2}. The gap between the `` + $MS_{i}^{2}$ ''  and the baseline demonstrates the effectiveness of the subtraction unit (SU).  It can be seen that the usage of SU has a significant improvement on the CVC-ColonDB dataset compared to the baseline, with the gain of 7.8\%, 7.4\%, 6.7\% and 4.4\% in terms of mDice, mIoU, $F_\beta^w$, and $E_\phi^{max}$, respectively. Next, we gradually add  $MS_{i}^{3}$, $MS_{i}^{4}$ and $MS_{i}^{5}$ to achieve multi-scale aggregation. The gap between the `` + $MS_{i}^{5}$ ''  and the `` + $MS_{i}^{2}$ '' quantitatively demonstrates the effectiveness of multi-scale strategy. To more intuitively show its effectiveness, we visualize features of each encoder level ($MS^i_1$) and the complementarity enhanced features ($CE^i$) in Fig~\ref{fig:ablation_study_visual_encoder_transition}. We can see that the multi-scale subtraction module can clearly highlight the difference between high-level features and other level features and propagate its localization effect to the low-level ones. Thus, both the global structural information and local boundary information is well depicted in the enhanced features of different levels.  Finally, we evaluate the benefit of $\mathcal{L}_{f}$.  Compared to the  `` + $MS_{i}^{5}$ '' model,  the  `` + $\mathcal{L}_{f}$ '' achieves significant performance improvement on the ETIS dataset, with the gain of 11.8\%, 14.1\%, 13.0\% and 5.5\% in terms of mDice, mIoU, $F_\beta^w$, and $E_\phi^{max}$, respectively. Besides,  we replace all subtraction units with the element-wise addition units and compare their performance. It can be seen that our subtraction units have significant advantage and no additional parameters are introduced.

\section{Discussion}
\textbf{Multi-scale Subtraction Module}: Different from previous addition operation, using subtraction in multi-scale module make resulted features input to the decoder have much less redundancy among different levels and their scale-specific properties are significantly enhanced. This mechanism can be explored in more segmentation tasks in the future.
\\
\textbf{LossNet}: LossNet is similar in form to perception loss~\cite{Ploss} that has been applied in many tasks, such as style transfer and inpainting.  While in those vision tasks, the perception-like loss is mainly used to speed the convergence of GAN and obtain high frequency information and ease checkerboard artifacts, but it does not bring obvious accuracy improvement. In our paper, the inputs are binary segmentation masks, LossNet can directly target the geometric features of the lesion and perform joint supervisions from the contour to the body, thereby improving the overall segmentation accuracy.
In the binary segmentation task, our work is the first one.
\section{Conclusion}
In this paper, we present a novel multi-scale subtraction network (MSNet) to automatically segment polyps from colonoscopy images.
We pyramidally concatenate multiple subtraction units to extract lower-order and higher-order cross-level complementary information and combine with level-specific information to enhance multi-scale feature representation.
%
Besides, we design a loss function based on a training-free network to supervise the prediction from different feature levels, which can optimize the segmentation on both structure and details during the backward phase.  
Extensive experimental results demonstrate that MSNet notably outperforms the state-of-the-art methods under different evaluation metrics. Moreover, the proposed model runs at the fastest speed of $\sim$70fps among the existing polyp segmentation methods.
\\
\noindent{\textbf{Acknowledgements.}}
This work was supported in part by the National Natural Science Foundation of China \#61876202, \#61725202, \#61751212 and \#61829102,
the Dalian Science and Technology Innovation Foundation \#2019J12GX039, 
and the Fundamental Research Funds for the Central Universities \#DUT20ZD212. 

%
%
%
\bibliographystyle{paper305}
\bibliography{paper305}

\end{document}